\pdfoutput=1
\documentclass[11pt]{article}

\usepackage[]{acl}

\usepackage{times}
\usepackage{latexsym}
\usepackage{bm}
\usepackage{amsmath}
\usepackage{amssymb}
\usepackage{color}
\usepackage{subcaption}
\usepackage{graphicx}
\usepackage{booktabs}
\usepackage{colortbl}
\usepackage{stmaryrd}
\usepackage{natbib}
\usepackage{multirow}
\usepackage{CJKutf8}
\usepackage{ascii}

\usepackage[T1]{fontenc}
\usepackage[utf8]{inputenc}

\usepackage{microtype}

\title{MSP: Multi-Stage Prompting for Making \\ Pre-trained Language Models Better Translators}

\author{
    Zhixing Tan$^{1,3, 4}$, Xiangwen Zhang$^{6}$, Shuo Wang$^{1,3,4}$, \and Yang Liu$^{1,2,3,4,5}$ \\
    $^1$Department of Computer Science and Technology, Tsinghua University, Beijing, China \\
    $^2$Institute for AI Industry Research, Tsinghua University, Beijing, China \\
    $^3$Institute for Artificial Intelligence, Tsinghua University, Beijing, China \\
    $^4$Beijing National Research Center for Information Science and Technology \\
    $^5$International Innovation Center of Tsinghua University, Shanghai, China \\
    $^6$Kuaishou Tech, Co.
}

\begin{document}
\maketitle

{\let\thefootnote\relax\footnotetext{Corresponding to: Z. Tan ({\asciifamily zxtan@tsinghua.edu.cn}) and Y. Liu ({\asciifamily liuyang2011@tsinghua.edu.cn})}}

\begin{abstract}
Prompting has recently been shown as a promising approach for applying pre-trained language models to perform downstream tasks. We present \underline{M}ulti-\underline{S}tage \underline{P}rompting, a simple and automatic approach for leveraging pre-trained language models to translation tasks. To better mitigate the discrepancy between pre-training and translation, MSP divides the translation process via pre-trained language models into 
multiple separate stages: the encoding stage, the re-encoding stage, and the decoding stage. During each stage, we independently apply different continuous prompts for allowing pre-trained language models better shift to translation tasks. We conduct extensive experiments on three translation tasks. Experiments show that our method can significantly improve the translation performance of pre-trained language models.~\footnote{Source code is available at \url{https://github.com/THUNLP-MT/PLM4MT}.}
\end{abstract}

\section{Introduction}

\begin{figure*}[th]
\centering 
    \begin{subfigure}[t]{0.35\textwidth}
        \centering
        \includegraphics[width=\textwidth]{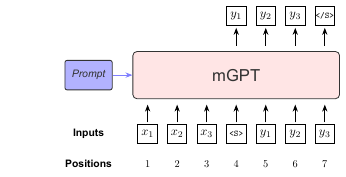}
        \caption{Basic (single-stage) prompting for MT.}
    \end{subfigure}
    \hspace{4em}
    \begin{subfigure}[t]{0.455\textwidth}
        \centering
        \includegraphics[width=\textwidth]{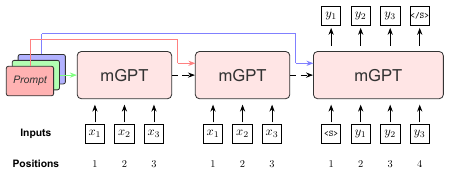}
        \caption{Multi-stage prompting.}
    \end{subfigure}
    \caption{Overview of using prompts for steering a multilingual GPT (mGPT) model to machine translation tasks. Note that we reset the position ids during each stage in multi-stage prompting for ease of implementation. All stages use the same mGPT model.}\label{fig:overview}
\end{figure*}

Prompting~\cite{brown2020gpt3,lester2021power}, which refers to the approach of generating task-specific outputs from language models (LMs) by conditioning on extra information (known as \emph{prompts}), has emerged as a new way of using LMs to perform natural language processing (NLP) tasks~\cite{gao2020making,liu2021pre}. While being efficient in parameters~\cite{lester2021power}, prompting can enable mixed-task inference, which is not possible for other related approaches like finetuning or adapter-based tuning~\cite{li2021prefix,lester2021power}. Prompting also opens the possibility of using a single pre-trained LM to perform all NLP tasks~\cite{liu2021pre}.

Machine translation (MT), which involves transformations between two languages, is considered one of the most challenging tasks in NLP~\cite{koehn2017six}. While neural machine translation (NMT)~\cite{sutskever2014sequence,bahdanau2015nmt,vaswani2017attention} is the current \emph{de facto} approach for machine translation, using pre-trained LMs as translators via prompting is appealing in several aspects. For example, for the method described in this paper, supporting a new translation direction with a pre-trained LM occupies disk spaces below 20M, which is much smaller than training a separate neural machine translation model, where the model size is typically larger than 60M per language pair for the Transformer architecture.~\footnote{Assume using the \emph{transformer-base} setting with a vocabulary size of 32K.} Furthermore, the pre-trained LM also retains the ability to perform other downstream tasks, which is an important characteristic that has not been validated available on neural machine translation models.

However, it is challenging to leverage pre-trained LMs to translation tasks via prompting. First, finding an appropriate prompt for a translation task is not trivial and requires specific designs~\cite{brown2020gpt3,gao2020making,li2021prefix,lester2021power}. Second, the prompting method with a single prompt may be sub-optimal for steering pre-trained LMs to translation tasks, as there is a clear discrepancy between the objectives of translation and pre-training. Translation imposes strict \emph{semantic equivalence} and \emph{language space} constraint, in which a source sentence must translate to a semantically equivalent sentence in the target language space. As the objective of pre-training is usually to reconstruct parts of the input sentence~\cite{radford2018improving,devlin2019bert}, the generation of a pre-trained LM conditioned on a source sentence will likely be in the source language space with non-equivalent semantics. Therefore, using a single prompt to guide the LM for mitigating both the semantic and language gap is likely to be sub-optimal. Third, prevalent generative LMs such as GPTs use a decoder-only architecture~\cite{radford2018improving}, which is unidirectional and may be sub-optimal for encoding source sentences~\cite{devlin2019bert}. While recent works in prompting like prefix-tuning~\cite{li2021prefix} or prompt tuning~\cite{lester2021power} alleviate the first challenge by introducing differentiable continuous prompts, the last two challenges remain to be addressed.

In this paper, we present Multi-Stage Prompting (MSP) for addressing the challenges of steering pre-trained language models to translation tasks. MSP encapsulates the idea of breaking translation tasks into simpler consecutive stages, allowing the pre-trained LM to learn ``smoother transitions'' to translation tasks by providing different prompts at different stages. For GPT-style pre-trained LMs, we design a three-stage prompting scheme for modeling the translation process, which consists of an \emph{encoding stage}, a \emph{re-encoding stage}, and a \emph{decoding stage}. Specifically, the pre-trained LM focuses on learning source representations at the encoding stage and learns refined bidirectional representations by re-encoding source sentences at the re-encoding stage. Therefore, the LM can produce better translations with refined source representations at the decoding stage. Following prefix-tuning~\cite{li2021prefix} and prompt tuning~\cite{lester2021power}, we use independent trainable continuous prompts at different stages, which are learned through back-propagation. The difference between basic (single-stage) prompting and multi-stage prompting is illustrated in Figure~\ref{fig:overview}. 

We demonstrate the effectiveness of our method with a multilingual GPT (mGPT) model on Romanian-English, English-German, and English-Chinese translation tasks. Experiments verify that compared with prompt tuning or prefix-tuning, MSP can significantly improve the translation performance of pre-trained LMs. Our method improves the translation performance of pre-trained language models via prompt tuning and prefix-tuning by 18.6 and 4.1 BLEU points on average over the three translation tasks, respectively, suggesting that MSP is a more effective prompting method for translation tasks.

\section{Background}

\subsection{Prompting}
Prompting is an approach of using an LM to perform downstream tasks by adding extra information for the LM to condition during its generation~\cite{lester2021power}. This extra information, also known as a \emph{prompt}, plays an important role in prompting methods and is often prepended to LM's input for better control of its generation. Depending on the form of prompts, prompting methods can be divided into two categories: using textual prompts or using continuous prompts.

Textual prompts are typically composed of natural language tokens. As a representative approach of textual prompts, \citet{brown2020gpt3} use manually designed prompts to steer GPT-3's generation. A typical prompt used in GPT-3 consists of a task description and a few task-specific examples. \citet{gao2020making} and \citet{shin2020autoprompt} propose different automatic methods to generate textual prompts. Textual prompts are typically understandable by humans. However, \citet{shin2020autoprompt} indicate that automatically generated textual prompts may lack interpretability. 

Continuous prompts, which consist of a sequence of continuous vectors, have gained increasing popularity recently. For example, in \cite{li2021prefix}, the continuous prompts consist of a sequence of key-value pairs (also called prefixes). \citet{lester2021power} propose a simplified version of continuous prompts, which consists of virtual tokens that are only added to the embedding layer. Compared with textual prompts, using continuous prompts is generally more powerful but less interpretable~\cite{lester2021power}.

\subsection{mGPT}
In this paper, we use GPT~\cite{radford2018improving,radford2019language,brown2020gpt3} as the backbone LM for machine translation tasks. GPTs are a series of causal language models based on the Transformer architecture~\cite{vaswani2017attention}. To be more suitable for translation tasks that involve multiple languages, we introduce a \emph{multilingual GPT} (mGPT) model instead of using a standard GPT-2 model.~\footnote{We release our checkpoint at \url{https://huggingface.co/THUMT/mGPT}.} The main difference between mGPT and GPT-2 is the training data. mGPT is trained on the mC4 dataset~\cite{xue2020mt5}, which is a multilingual dataset covering over 101 languages. For further details about mGPT, please refer to Appendix~\ref{sec:appendix_mgpt}.

Let $\mathbf{z}=[z_1, \ldots, z_n]$ be a sequence of tokens, mGPT uses an autoregressive Transformer network to model the conditional probability $P(z_{t}|\mathbf{z}_{<t})$, where $t \in [1,n]$ and $\mathbf{z}_{<t}=[z_1,\ldots,z_{t-1}]$. We use $f_{\mathrm{LM}}(\bm{z}, \bm{H};\bm{\theta})$ to denote the Transformer network, where $\bm{z}$ is a word embedding, $\bm{H}$ is a sequence of past activations, and $\bm{\theta}$ denotes the parameters of the Transformer network. 

Initially, the inputs to the Transformer network are $z_1$ and $\bm{H}_0$, where $\bm{H}_0$ is an empty sequence. The Transformer network produces two outputs: the final output $\bm{g}_1 \in \mathbb{R}^{d}$ and the activation $\bm{h}_1 \in \mathbb{R}^{2N \times d}$,~\footnote{$\bm{h}$ is a concatenation of a set of key-value pairs $\{\langle \bm{k}^{(i)}, \bm{v}^{(i)} \rangle | i = 1\ldots N \}$ in the Transformer network.} where $d$ denotes the hidden size of the Transformer network and $N$ is the number of layers of the Transformer network.

For subsequent inputs $z_t$ and $\bm{H}_{t-1}$, where $\bm{H}_{t-1}=[\bm{h}_1,\ldots,\bm{h}_{t-1}]$, the computation is formally described as
\begin{align}
\bm{g}_t, \bm{h}_t = f_{\mathrm{LM}}(\bm{e}_{z_t}, \bm{H}_{t-1}),\label{eq:plm}
\end{align}
where $\bm{e}_{z_t}$ denotes the word embedding of $z_t$. To make the notation simpler, we use the following equation to denote the repeated application of $f_{\mathrm{LM}}$ over a sequence $\mathbf{z}^{i:j} = [z_i,\ldots,z_j]$ given past activations $\bm{A}$:
\begin{align}
\bm{G}^{i:j}, \bm{H}^{i:j} = f_{\mathrm{LM}}(\bm{Z}^{i:j}, \bm{A}),
\end{align}
where $\bm{Z}^{i:j}=[\bm{e}_{z_i},\ldots,\bm{e}_{z_j}]$, $\bm{G}^{i:j}=[\bm{g}_{i},\ldots,\bm{g}_{j}]$, and $\bm{H}^{i:j}=[\bm{h}_{i},\ldots,\bm{h}_{j}]$. 

Finally, the conditional probability $P(z_{t}|\mathbf{z}_{<t})$ is modeled as follows:
\begin{align}
P(z_{t}|\mathbf{z}_{<t}) = \frac{\exp{(\bm{e}_{z_t}^{\mathsf{T}} \cdot \bm{g}_t})}{\sum_{i=1}^{|V|} \exp{(\bm{e}_{z_i}^{\mathsf{T}} \cdot \bm{g}_t})},
\end{align}
where $|V|$ is the vocabulary size, and ``$\cdot$'' denotes matrix production.

\section{Multi-Stage Prompting}

\begin{figure}[t]
\centering 
    \includegraphics[width=0.3\textwidth]{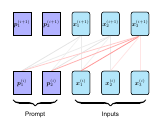}
    \caption{A deep continuous prompt is prepended to the inputs in all attention layers, which affects the computation of all attention layers. We do not distinguish keys and values here for simplicity.}\label{fig:att}
\end{figure}

\begin{figure*}[th]
\centering 
    \includegraphics[width=0.8\textwidth]{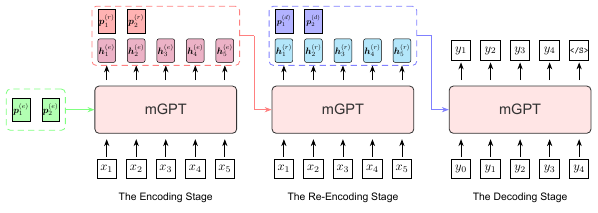}
    \caption{Detailed computations involved in the multi-stage prompting for machine translation tasks. We use rectangles to denote prompt vectors and rounded rectangles to denote activations.}\label{fig:msp}
\end{figure*}

We propose multi-stage prompting (MSP), a simple and lightweight method for steering pre-trained LMs to translation tasks. We first describe the concept of deep continuous prompts in Section~\ref{sec:dsp}. Then we detail the stages and training objective in Section~\ref{sec:msp} and Section~\ref{sec:obj}, respectively. Finally, we describe the reparameterization of deep continuous prompts in Section~\ref{sec:reparam}.

\subsection{Deep Continuous Prompts}\label{sec:dsp}
We adopt ``continuous prompts''~\cite{li2021prefix,lester2021power} instead of using textual prompts in our method. Using continuous prompts allows learning through differentiable methods like back-propagation~\cite{lester2021power}. To be specific, we use deep continuous prompts which are in the same form as in \cite{li2021prefix}. Formally, a prompt $\bm{P}$ is a sequence of $L$ continuous vectors $[\bm{p}_1,\ldots,\bm{p}_L]$. Each vector $\bm{p}_i \ (1 \le i \le L)$ is a concatenation of key-value pairs in all $N$ Transformer layers, which directly affect the computation of every attention layer. Therefore, the dimension of $\bm{p}_i$ is $2N\times d$. We give an illustration of conditioning on a deep continuous prompt in Figure~\ref{fig:att}. 

\subsection{Stages}\label{sec:msp}
To effectively mitigate the semantic and language gap between the pre-training and translation, we propose multi-stage prompting which divides the procedure of using pre-trained LMs as translators into three separate stages: the encoding, the re-encoding, and the decoding stages. Given different prompts at different stages, the pre-trained LM is expected to behave differently during each stage and is more capable of generating translations.

Given a source sentence $\mathbf{x}=[x_1,\ldots,x_S]$ and a target sentence $\mathbf{y}=[y_1,\ldots,y_T]$, the details of the three stages are described as follows:

\paragraph{The Encoding Stage.} At the encoding stage, the pre-trained LM encodes the source sentence $\mathbf{x}$ into a sequence of activations $\bm{H}_{e}^{1:S}$ by using an encoding stage prompt $\bm{P}_{e}$. This procedure is the same as basic prompting. Formally, it can be described as follows:
\begin{align}
\bm{G}_{e}^{1:S},\bm{H}_{e}^{1:S} = f_{\mathrm{LM}}(\bm{X}^{1:S}, \bm{P}_{e}).
\end{align}

\paragraph{The Re-encoding Stage.} At the re-encoding stage, the pre-trained LM produces fine-grained representations of the source sentence by re-encoding $\mathbf{x}$ given past activations $\bm{H}_e^{1:S}$ and a re-encoding stage prompt $\bm{P}_r$, which allows each representation to condition on all words in $\mathbf{x}$. This procedure can be described as
\begin{align}
\bm{G}_{r}^{1:S},\bm{H}_{r}^{1:S} = f_{\mathrm{LM}}(\bm{X}^{1:S}, \llbracket \bm{P}_{r}; \bm{H}_{e}^{1:S} \rrbracket),
\end{align}
where $\llbracket \bm{P}_r; \bm{H}_e^{1:S} \rrbracket$ denotes the concatenation of two sequences $\bm{P}_r$ and $\bm{H}_e^{1:S}$. It is also possible to employ more than one re-encoding stage, allowing the pre-trained LM to obtain further refined representations of the source sentence.

\paragraph{The Decoding Stage.} Finally, we obtain the hidden vectors $\bm{G}_d^{1:T}$ for predicting the probability of the target sentence $\mathbf{y}$ at the decoding stage, given the refined source representations $\bm{H}_r^{1:S}$ and a decoding stage prompt $\bm{P}_d$:
\begin{align}
\bm{G}_d^{1:T},\bm{H}_d^{1:T} = f_{\mathrm{LM}}(\bm{Y}^{1:T}, \llbracket \bm{P}_d; \bm{H}_r^{1:S} \rrbracket). \label{eq:dec}
\end{align}

Figure~\ref{fig:msp} gives a detailed illustration of MSP. By dividing the translation process into multiple stages and applying different prompts, we expect the pre-trained LM model can generate better translations.

\subsection{Training Objective}\label{sec:obj}
We use the cross-entropy loss for learning prompts. Given $\bm{G}_d^{1:T}=[\bm{g}_1^{(d)},\ldots,\bm{g}_T^{(d)}]$ in Eq.~(\ref{eq:dec}), the training objective is formally described as follows:
\begin{align}
\begin{split}
\mathcal{L} &= -\frac{1}{T}\sum_{t=1}^T \log P(y_t|\mathbf{y}_{<t},\mathbf{x}) \\
            &= -\frac{1}{T}\sum_{t=1}^T \log \frac{\exp{(\bm{e}_{z_t}^{\mathsf{T}} \cdot \bm{g}_t^{(d)}})}{\sum_{i=1}^{|V|} \exp{(\bm{e}_{z_i}^{\mathsf{T}} \cdot \bm{g}_t^{(d)}})}.
\end{split}
\end{align}
Note that the parameters $\bm{\theta}$ of the pre-trained LM are fixed during training.

\subsection{Reparameterization}\label{sec:reparam}

\citet{li2021prefix} suggest that using a neural network to reparameterize continuous prompts is more robust to different choices of hyperparameters. In contrast to their approach which uses an MLP network to reparameterize continuous prompts, we introduce a much simpler \emph{scaled reparameterization} method, in which a continuous prompt is reparameterized as a product of a learnable scalar and an embedding. More precisely, the reparameterization of the three prompts are as follows:
\begin{align}
\bm{P}_e &= \max(\alpha_e, 1.0) \times \bm{\phi}_e, \\
\bm{P}_r &= \max(\alpha_r, 1.0) \times \bm{\phi}_r, \\
\bm{P}_d &= \max(\alpha_d, 1.0) \times \bm{\phi}_d,
\end{align}
where $\bm{\phi}_e \in \mathbb{R}^{2N \times d}$, $\bm{\phi}_r \in \mathbb{R}^{2N \times d}$, and $\bm{\phi}_d \in \mathbb{R}^{2N \times d}$. $\alpha_e$, $\alpha_r$, and $\alpha_d$ are initialized to 1.0 at the beginning of training. Therefore, the set of trainable parameters $\bm{\phi}$ in our method is $\bm{\phi} = \{ \alpha_e, \alpha_r, \alpha_d, \bm{\phi}_e, \bm{\phi}_r, \bm{\phi}_d \}$, which contains much less tunable parameters than an MLP network. 

Scaled reparameterization enables directly adjusting the value of prompts by a tunable scaling factor, leading to a much faster convergence without loss of performance. Further analysis is presented in Section~\ref{sec:rep}.

\section{Experiments}

\begin{table*}[th]
\begin{center}
\begin{tabular}{lrrrrr}\toprule
\textbf{Method} & \textbf{\#Params.} & \textbf{Ro-En} & \textbf{En-De} & \textbf{En-Zh} & \textbf{Average} \\\midrule
Prompt Tuning & 131K &  17.7 & 5.9 & 4.5 & 9.4 \\
Prefix-Tuning & 26M & 32.5 & 17.5 & 21.9 & 23.9 \\
MSP (\emph{Ours}) & 19M & \textbf{34.7} & \textbf{21.2} & \textbf{28.1} & \textbf{28.0} \\
\bottomrule
\end{tabular}
\caption{BLEU score on three different translation tasks for different prompting methods. All prompting methods use the same pre-trained language model ``mGPT''. ``\#Params.'' denotes the number of tunable parameters during training.}\label{tab:main}
\end{center}
\end{table*}

\subsection{Setup}

\paragraph{Datasets} We conduct experiments on Romanian-English (Ro-En), English-German (En-De), and English-Chinese (En-Zh) translation tasks to verify our proposed method. For the Ro-En translation task, we used the WMT16 Romanian-English dataset, which consists of 0.6M bilingual sentence pairs and 2M back-translated sentence pairs.\footnote{
\url{http://data.statmt.org/rsennrich/wmt16_backtranslations/ro-en}} We used \textit{newsdev2016} as the development set and \textit{newstest2016} as the test set. For the En-De translation task, we used the WMT14 English-German dataset, which consists of 4.5M sentence pairs. The development set is \textit{newstest2013} and the test set is \textit{newstest2014}. For the En-Zh translation task, we used the WMT20 English-Chinese dataset as the training corpus, which consists of 28M sentence pairs. The development set is \textit{newstest2019} and the test set is \textit{newstest2020}. The details of preprocessing and postprocessing are given in Appendix~\ref{sec:appendix_proc}.

\paragraph{Metric.}
We used case-sensitive BLEU~\cite{papineni2002bleu} as the evaluation metric. The BLEU score is calculated using the \textsc{SacreBLEU} toolkit~\cite{post2018call}.\footnote{Signature: nrefs:1|case:mixed|eff:no|tok:\{13a,zh\}|\\smooth:exp|version:2.0.0}

\paragraph{Baselines.} We used the mGPT model as the backbone LM in all our experiments, which contains 560M parameters. We compare our method with the following prompting methods:~\footnote{In our preliminary experiments, we also experimented with the few-shot approach as described in ~\cite{brown2020gpt3}. However, we found mGPT often failed to generate meaningful translations.}

\begin{itemize}
\item Prompt tuning~\cite{lester2021power}. A prompting method that only prepends virtual tokens to the embedding layer of pre-trained LMs.
\item Prefix-tuning~\cite{li2021prefix}. A prompting method that uses deep continuous prompts, which prepend virtual tokens to all key-value pairs in attention layers of pre-trained LMs. We use an MLP network to reparameterize a continuous prompt during training as suggested in \cite{li2021prefix}.
\end{itemize}

% Training speed.
\paragraph{Implementations.} All our models are trained on a machine with 8 RTX 3090Ti GPUs. For all prompting methods, we set the prompt length to 128. For the training, we use the Glorot uniform initilalizer~\cite{glorot2010understanding} to initialize tunable parameters unless otherwise noted. We use Adam~\cite{kingma2014adam} ($\beta_1$ = 0.9, $\beta_2$ = 0.98 and $\epsilon$ = 1$\times$ $10^{-9}$) as the optimizer with a batch size of roughly 32K tokens. We use the same learning rate schedule as described in \cite{vaswani2017attention}. The number of warmup steps is set to 4K. We set the maximum learning rate to 0.02 for prompt tuning and MSP, and 7e-4 for prefix-tuning.\footnote{We found using a large learning rate for prefix-tuning would result in unstable training.} We train prompts for a total of 80K steps for prompt tuning and prefix-tuning, and 40K steps for MSP. For the inference, we use the beam search algorithm to obtain translation from the mGPT model, and the beam size is set to 4. The length penalty is determined by the results evaluated on the development set. We set the length penalty to 1.0 for the En-Zh translation task and 0.0 for other translation tasks. We implement our models on top of the THUMT~\cite{tan2020thumt} toolkit and the Transformers library~\cite{wolf2020transformers}.

\subsection{Main Results}

\begin{table*}[th]
\centering
\begin{tabular}{l c c c c}
\toprule
\textbf{LM} & \textbf{Architecture} & \textbf{\#M-Params.} & \textbf{Method} & \textbf{BLEU} \\\midrule
%Transformer & Encoder-Decoder & 432M & - & 41.8 \\\midrule 
mT5-XXL~\cite{zhang2021cpm} & Encoder-Decoder & 13B & Finetuning & 24.0 \\
CPM-2~\cite{zhang2021cpm} & Encoder-Decoder & 11B & Prompt Tuning & 24.1 \\
CPM-2~\cite{zhang2021cpm} & Encoder-Decoder & 11B & Finetuning & 26.2 \\
Ernie 3.0~\cite{sun2021ernie} & Encoder-Decoder & 10B & Finetuning & 26.8 \\\midrule
mGPT (\emph{Ours}) & Decoder & 560M & MSP & \textbf{28.1} \\\bottomrule
\end{tabular}
\caption{Comparisons with previous studies on the WMT20 En-Zh translation task. ``\#M-Params.'' indicates the number of parameters of pre-trained LMs.}\label{tab:enzh}
\end{table*}

Table~\ref{tab:main} shows the results for the Ro-En, En-De, and En-Zh translation tasks. 

As the most parameter-efficient among the three prompting methods, prompt tuning introduces only 131K parameters during training for each translation task. However, it only achieves 9.4 BLEU points on average over the three translation tasks. \citet{lester2021power} indicate that language model capacity is a key ingredient for prompt tuning to succeed. As mGPT is a pre-trained LM with only 560M parameters, the results coincide with the conclusion of \citet{lester2021power}.

Prefix-tuning, which uses deep continuous prompts, achieves an average of 23.9 BLEU points over the three translation tasks. The results indicate that using deep continuous prompts is beneficial for steering mGPT to translation tasks. However, introducing deep continuous prompts inevitably requires more free parameters. The MLP network used in prefix-tuning introduces about 26M parameters for each translation task during training in our experiments.

Finally, MSP achieves 28.0 BLEU points on average over the three translation directions and outperforms prompt tuning and prefix-tuning by 18.6 and 4.1 BLEU points, respectively. MSP introduces 19M parameters for each translation task during training, which is more than prompt tuning but less than prefix-tuning. MSP explicitly divides the translation process using mGPT into separate stages, which are not present in prompt tuning and prefix-tuning. The results suggest that MSP is more effective in instructing pre-trained LMs to perform translation than prompt tuning and prefix-tuning. 

\subsection{Comparison with Other LMs}

Table~\ref{tab:enzh} gives the results of mT5-XXL~\cite{zhang2021cpm}, CPM-2~\cite{zhang2021cpm}, Ernie 3.0~\cite{sun2021ernie}, and mGPT on the WMT20 En-Zh translation task. Except for mGPT, other LMs are based on the encoder-decoder architecture. Despite using a much smaller pre-trained LM with about 5\% parameters of mT5-XXL, CPM-2, and Ernie 3.0, MSP achieves the best performance on the En-Zh translation task. Therefore, we show that MSP is an efficient and effective approach to steering pre-trained LMs to translation tasks.

\subsection{Comparison with Transformer}

\begin{table*}[th]
\begin{center}
\begin{tabular}{lccccccc}\toprule
\textbf{Model} & \textbf{\#Params.} & \textbf{Bg} & \textbf{Es} & \textbf{It} & \textbf{Ru} & \textbf{Tr} & \textbf{Avg.} \\\hline
\multicolumn{8}{c}{\textit{X$\rightarrow$En}} \\\hline
Transformer & 437M & 35.2 & 38.0 & 34.2 & 22.6 & 21.0 & 30.2 \\
mGPT (MSP) & 19M & \textbf{38.9} & \textbf{42.1} & \textbf{37.8} & \textbf{24.4} & \textbf{24.9} & \textbf{33.6} \\\hline
\multicolumn{8}{c}{\textit{En$\rightarrow$X}} \\\hline
Transformer & 437M & 29.2 & 34.0 & 29.2 & 16.7 & 11.6 & 24.1 \\
mGPT (MSP) & 19M & \textbf{34.1} & \textbf{38.4} & \textbf{32.8} & \textbf{19.2} & \textbf{15.6} & \textbf{28.0} \\
\bottomrule
\end{tabular}
\caption{Results on the TedTalks ``X$\rightarrow$En'' and ``En$\rightarrow$X'' translation directions. For MSP, each translation direction introduces 19M parameters.}\label{tab:result_tedtalks}
\end{center}
\end{table*}

\begin{table}[th]
\centering
\resizebox{0.4\textwidth}{!}{ 
\begin{tabular}{l c c}
\toprule
\textbf{Model} & \textbf{\#Params.} & \textbf{BLEU} \\\midrule
Transformer (big) & 450M & 27.9 \\
mGPT (MSP) & 19M & 21.2 \\
\bottomrule
\end{tabular}}
\caption{Results on the WMT14 En-De dataset. ``\#Params.'' denotes the number of tunable parameters during training.}\label{tab:result_ende}
\end{table}

We compare our method with the state-of-the-art Transformer NMT model~\cite{vaswani2017attention}~\footnote{We used the \emph{transformer-big} setting. Tokenizations and vocabularies are the same with mGPT for fair comparisons.} on the TedTalks dataset~\cite{blackwood2018multilingual} and the WMT14 English-German dataset. TedTalks dataset is an English-centric multilingual corpus including 59 languages with around 3K to 200K sentence pairs per language pair. For the sake of simplicity, we only report results for 5 selected languages that contain more than 150K sentence pairs. However, the Transformer model is trained on all available parallel sentences covering 59 languages, serving as a strong NMT baseline. For mGPT with MSP, we individually train the model on each language pair following the same procedure described in this paper.

The results of ``X$\rightarrow$En'' and ``En$\rightarrow$X'' directions are shown in Table~\ref{tab:result_tedtalks}. Although mGPT with MSP is independently trained on each language pair, the model still outperforms the strong multilingual NMT baseline by 3.4 and 3.9 BLEU points on ``X-En'' and ``En-X'' directions, respectively. The results demonstrate that using pre-trained LMs as translators with an appropriate prompting method has the potential to excel a strong Transformer NMT model.

Table~\ref{tab:result_ende} shows the comparison between Transformer and our mGPT model with MSP on the En-De translation task. While there is still a noticeable performance gap between Transformer and mGPT with MSP, using mGPT as a translator with MSP is much more parameter-efficient than training a separate NMT model. Supporting En-De translation with mGPT only introduces 19M parameters with MSP method. In comparison, the model size of the Transformer model for En-De translation is 450M. While mGPT model can perform other downstream tasks by providing different prompts, such abilities have not been validated on the Transformer NMT model. Besides being efficient in disk spaces, learning prompts for the En-De translation task are also faster than training a separate NMT model. It takes 21 hours to train prompts for MSP, whereas 72 hours for training a Transformer model.

\subsection{Effect of Prompt Length}

\begin{figure}[!t]{}
\centering
\includegraphics[width=0.4\textwidth]{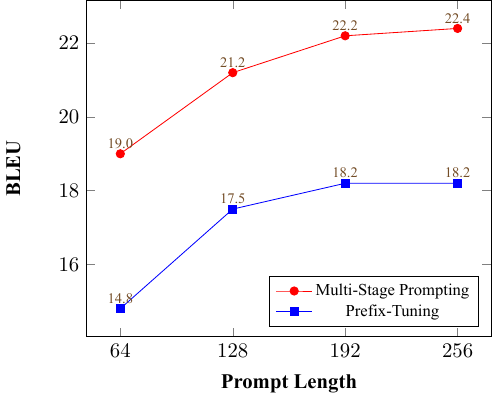}
\caption{Comparison between MSP and prefix-tuning on the WMT14 En-De translation task with different prompt lengths.}\label{fig:pl}
\end{figure}

Figure~\ref{fig:pl} shows the effect of prompt length for prefix-tuning and MSP. We omit the comparison to prompt tuning because of its inferior performance. We found that using longer prompts generally leads to better performance for both prefix-tuning and MSP, but with diminishing returns. This finding is consistent with previous studies~\cite{li2021prefix,lester2021power}. Furthermore, MSP consistently outperforms prefix-tuning when using the same prompt length. Even MSP with a prompt length of 64 performs better than prefix-tuning with a prompt length of 256 (19.0 vs. 18.2). The results further confirm that MSP is a better prompting method than prefix-tuning for steering pre-trained LMs to translation tasks. For the inference time, we found longer prompts do not significantly affect the decoding speed on GPUs as the computation of attention layers are highly parallel, which is also consistent with the findings of \citet{li2021prefix}.

\subsection{Effect of Stages}\label{sec:stages}

\begin{table*}[th]
\centering
\begin{tabular}{l r c c r r}
\toprule
\textbf{Method} & \textbf{\#Params.} & \textbf{Training} & \textbf{Inference} & \textbf{En-De} & \textbf{En-Zh} \\\midrule
Single-stage & 6.3M & 14h & 0.10 s/sent. & 17.9 & 22.8 \\
Two-stage (encoding/decoding) & 12.6M & 14h & 0.10 s/sent. & 20.2 & 25.2 \\
\quad $+$ Re-encoding (\emph{default}) & 19.0M & 21h & 0.11 s/sent. & 21.2 & 28.1 \\
\quad\quad $+$ 2nd Re-encoding & 25.1M & 29h & 0.11 s/sent. & 21.8 & 28.4 \\ 
\quad\quad $+$ Prompt sharing & 6.3M & 21h & 0.11 s/sent. & 19.8 & 24.5 \\ 
\bottomrule
\end{tabular}
\caption{Comparison of using different stage settings on the WMT14 En-De translation task and WMT20 Zh-En translation task. ``\#Params.'' denotes the number of trainable parameters. ``Training'' denotes the total training time. ``Inference'' denotes the inference speed measured on the test set using 8 GPUs. ``s/sent.'' denotes seconds per sentence. All experiments use \emph{scaled reparameterization} for fair comparison.}\label{tab:stages}
\end{table*}

Table~\ref{tab:stages} shows the comparison of using different stage settings on the WMT14 En-De and the WMT20 En-Zh translation tasks. For single-stage prompting, we also adopt scaled reparameterization instead of MLP reparameterization for a fair comparison. On the WMT14 En-De translation task, using single-stage prompting achieves 17.9 BLEU points. By comparison, explicitly separating encoding and decoding stages improve the translation performance over single-stage prompting by 2.3 BLEU points, which indicates the importance of differentiating stages. Adding a re-encoding stage further improves the translation performance by 1.0 BLEU point, suggesting that the re-encoding stage is effective. Adding a second re-encoding stage further improves the translation performance by 0.6 BLEU points. Although adding stages introduces more trainable parameters, it should be noted that sharing a single prompt for the encoding/re-encoding/decoding stages also improves over the single-stage prompting by 1.9 BLEU points. The results suggest that most improvements are attributed to the explicit separation of stages rather than increased parameters. Adding more stages generally slows the training speed. However, we do not observe notable inference speed drop as re-encoding stages are computed one time in parallel during inference. On the En-Zh translation task, the results are consistent with the results on the En-De translation task. Therefore, we conclude that using more stages helps improve the translation quality.

\subsection{Effect of Reparameterization}\label{sec:rep}

\begin{figure}[!t]{}
\centering
\includegraphics[width=0.4\textwidth]{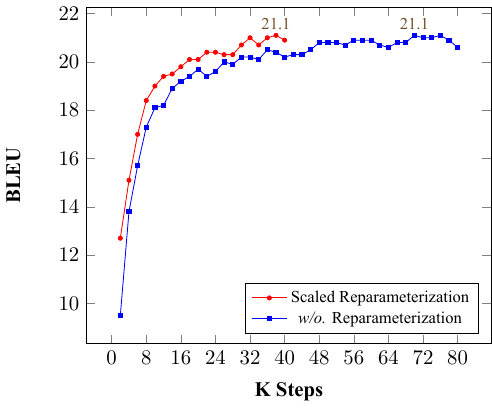}
\caption{Comparison between using scaled reparameterization and without using reparameterization on the WMT14 translation task. The BLEU score is evaluated on \emph{newstest2013}.}\label{fig:reparam}
\end{figure}

Figure~\ref{fig:reparam} shows the comparison between MSP using scaled reparameterization and without using reparameterization. Using scaled reparameterization converges faster than without using reparameterization. These two methods achieve nearly the same translation performance when the training is converged. As a result, using scaled reparameterization can make the convergence much faster and reduce the total training time.

\subsection{Analysis}

\begin{table}[t]
\centering
\resizebox{0.4\textwidth}{!}{ 
\begin{tabular}{l l}
\toprule
\textbf{Prompt} & \multicolumn{1}{c}{\textbf{Distribution}} \\\midrule
\emph{w/o} prompt & en~(16\%), ru~(10\%) \\
Prefix-tuning & zh~(80\%), ja~(12\%)  \\
MSP (encoding stage) & en~(51\%), la~(14\%) \\
MSP (re-encoding stage) &  en~(24\%), la~(17\%) \\
MSP (decoding stage) & zh~(91\%), ja~(9\%) \\
\bottomrule
\end{tabular}
}
\caption{Language distribution of the free generations using mGPT by conditioning on different prompts learned by different prompting methods on the WMT20 En-Zh dataset.}\label{tab:gen}
\end{table}

\paragraph{Knowledge.} As continuous prompts are learned using bilingual sentence pairs, an interesting question arises: Is the translation knowledge stored in the continuous prompts or the pre-trained LM? To answer this question, we discard the prompts and feed the mGPT model a concatenation of a parallel sentence pair as an input, and calculate the cosine similarities between the source and target hidden activations on each mGPT layer. We found that although the prompts are not given, the nearest pairs of tokens between the source and target language frequently turn out to coincide with bilingual alignments. This finding reveals to some extent that the translation knowledge mainly resides in the pre-trained LM instead of the learned continuous prompts, while the prompts play a role in guiding the model to perform translation during generation. Examples are given in Appendix~\ref{sec:appendix_alignments}.

\paragraph{Bottleneck.} We study the bottleneck of the current prompting method. We train a separate Transformer encoder and an adapter network that directly maps a source sentence into a deep continuous prompt, leaving the mGPT model only serving as a decoder. This model introduces 378M tunable parameters and achieves 25.9 BLEU points on the WMT14 En-De translation task. Compared with 21.2 BLEU points by MSP, the result shows that there is still room to advance the translation performance of pre-trained LM by improving the prompting method, such as using \emph{dynamic prompts}~\cite{liu2021pre} for each input sentence. However, as translation knowledge may come from the pre-trained LM, the translation performance may be bottlenecked by the capability of the backbone LM.

\paragraph{Interpretability.} We did not find our learned prompts to be interpretable, which agrees with the findings of \citet{shin2020autoprompt} and \citet{lester2021power}. However, we do observe prompts of different stages changing the behavior of mGPT significantly. Specifically, we sample 100 examples generated from mGPT by providing prompts of different stages learned on the English-Chinese translation task and identify the language ids of generated texts using the \verb|langid| toolkit. The top-2 identified language distributions of each generation are shown in Table~\ref{tab:gen}. Without providing prompts, mGPT generates a random sentence from a random language. By given continuous prompts learned by prefix-tuning, the mGPT mostly generates texts related to Chinese. For MSP, it is noticeable that there is a transition from English to Chinese. mGPT generates English-related text given the encoding stage prompt. The distribution of languages becomes smoother when providing the prompt at the re-encoding stage. Finally, mGPT generates Chinese texts dominantly given the decoding stage prompt. The results coincide with our intuition that MSP helps the pre-trained LM to learn ``smoother transitions'' to the translation task.

\section{Related Work}
\paragraph{Prompting.} \citet{brown2020gpt3} propose to use a task description and a few examples to adapt the GPT-3 model to downstream tasks, which is referred to as in-context learning. Their prompts are manually designed. \citet{gao2020making} present LM-BFF for automatic prompts generation. They use T5 model~\cite{raffel2019exploring} to generate templates for prompting pre-trained LMs. \citet{li2021prefix} propose prefix-tuning, which uses continuous vectors as prompts. These prompts are trained using task-specific data and optimized through back-propagation. \citet{lester2021power} propose prompt tuning, which is similar to prefix-tuning but with fewer trainable parameters. Our method is also based on prompting. We use continuous prompts for steering PLMs to translation tasks. Unlike \citet{li2021prefix} and \citet{lester2021power} who present general frameworks, our method is focused on improving the translation performance of pre-trained LMs.

\paragraph{Using Pre-trained Models as Translators.} \citet{stickland2021recipes} investigate using BART and mBART models for machine translation tasks, their approach relies on adapter networks and finetuning parts of pre-trained LMs. \citet{guo2020incorporating} build a non-autoregressive NMT model by using a source BERT model as the encoder and a target BERT as the decoder with adapter layers. \citet{sun2021multilingual} propose grafting a source BERT model and a target GPT model for translation tasks. \citet{bapna2019simple} propose using small adapter layers to adapt a base NMT model to new translation tasks. All these methods are adapter-based, which injects tunable modules into the pre-trained models. As a result, the pre-trained models lose the ability to perform mixed-task inference. Our approach is based on prompting, which only uses prompts for steering the pre-trained LMs to translation tasks. \citet{zhang2021cpm} investigate using prompt tuning for steering CPM-2 model to the WMT20 English-Chinese translation task. Furthermore, their approach applied to encoder-decoder architecture pre-trained LMs while ours applied to decoder-only pre-trained LMs.

\section{Conclusion}
We have presented multi-stage prompting, a method for making pre-trained language models better translators. Experiments show that with multi-stage prompting, pre-trained LMs can generate better translations, showing the potential of using pre-trained LMs for translation tasks.

\section*{Acknowledgements}
This work was supported by the National Key R\&D Program of China (No. 2018YFB1005103), the National Natural Science Foundation of China (No. 62006138, No. 61925601), Institute Guo Qiang at Tsinghua University, and Huawei Noah's Ark Lab. We thank Kehai Chen for the discussion of this work and all anonymous reviewers for their valuable comments and suggestions on this work.

\bibliography{anthology,custom}

\begin{thebibliography}{29}
\expandafter\ifx\csname natexlab\endcsname\relax\def\natexlab#1{#1}\fi

\bibitem[{Bahdanau et~al.(2015)Bahdanau, Cho, and Bengio}]{bahdanau2015nmt}
Dzmitry Bahdanau, KyungHyun Cho, and Yoshua Bengio. 2015.
\newblock \href {https://arxiv.org/abs/1409.0473} {Neural machine translation
  by jointly learning to align and translate}.
\newblock In \emph{Proceedings of ICLR}.

\bibitem[{Bapna and Firat(2019)}]{bapna2019simple}
Ankur Bapna and Orhan Firat. 2019.
\newblock \href {https://doi.org/10.18653/v1/D19-1165} {Simple, scalable
  adaptation for neural machine translation}.
\newblock In \emph{Proceedings of EMNLP}, pages 1538--1548.

\bibitem[{Blackwood et~al.(2018)Blackwood, Ballesteros, and
  Ward}]{blackwood2018multilingual}
Graeme Blackwood, Miguel Ballesteros, and Todd Ward. 2018.
\newblock \href {https://aclanthology.org/C18-1263} {Multilingual neural
  machine translation with task-specific attention}.
\newblock In \emph{Proceedings of COLING}, pages 3112--3122.

\bibitem[{Brown et~al.(2020)Brown, Mann, Ryder, Subbiah, Kaplan, Dhariwal,
  Neelakantan, Shyam, Sastry, Askell et~al.}]{brown2020gpt3}
Tom Brown, Benjamin Mann, Nick Ryder, Melanie Subbiah, Jared~D Kaplan, Prafulla
  Dhariwal, Arvind Neelakantan, Pranav Shyam, Girish Sastry, Amanda Askell,
  et~al. 2020.
\newblock \href
  {https://proceedings.neurips.cc/paper/2020/file/1457c0d6bfcb4967418bfb8ac142f64a-Paper.pdf}
  {Language models are few-shot learners}.
\newblock In \emph{Advances in Neural Information Processing Systems},
  volume~33, pages 1877--1901.

\bibitem[{Devlin et~al.(2019)Devlin, Chang, Lee, and
  Toutanova}]{devlin2019bert}
Jacob Devlin, Ming-Wei Chang, Kenton Lee, and Kristina Toutanova. 2019.
\newblock \href {https://doi.org/10.18653/v1/N19-1423} {{BERT}: Pre-training of
  deep bidirectional transformers for language understanding}.
\newblock In \emph{Proceedings of NAACL}, pages 4171--4186.

\bibitem[{Gao et~al.(2020)Gao, Fisch, and Chen}]{gao2020making}
Tianyu Gao, Adam Fisch, and Danqi Chen. 2020.
\newblock \href {https://doi.org/10.18653/v1/2021.acl-long.295} {Making
  pre-trained language models better few-shot learners}.
\newblock In \emph{Proceedings of ACL}, pages 3816--3830.

\bibitem[{Glorot and Bengio(2010)}]{glorot2010understanding}
Xavier Glorot and Yoshua Bengio. 2010.
\newblock \href {https://proceedings.mlr.press/v9/glorot10a.html}
  {Understanding the difficulty of training deep feedforward neural networks}.
\newblock In \emph{Proceedings of the thirteenth international conference on
  artificial intelligence and statistics}, volume~9, pages 249--256.

\bibitem[{Guo et~al.(2020)Guo, Zhang, Xu, Wei, Chen, and
  Chen}]{guo2020incorporating}
Junliang Guo, Zhirui Zhang, Linli Xu, Hao-Ran Wei, Boxing Chen, and Enhong
  Chen. 2020.
\newblock \href
  {https://proceedings.neurips.cc/paper/2020/file/7a6a74cbe87bc60030a4bd041dd47b78-Paper.pdf}
  {{Incorporating BERT into Parallel Sequence Decoding with Adapters}}.
\newblock In \emph{Advances in Neural Information Processing Systems},
  volume~33, pages 10843--10854.

\bibitem[{Kingma and Ba(2015)}]{kingma2014adam}
Diederik~P Kingma and Jimmy Ba. 2015.
\newblock \href {https://arxiv.org/abs/1412.6980} {Adam: A method for
  stochastic optimization}.
\newblock In \emph{Proceedings of ICLR}.

\bibitem[{Koehn and Knowles(2017)}]{koehn2017six}
Philipp Koehn and Rebecca Knowles. 2017.
\newblock \href {https://doi.org/10.18653/v1/W17-3204} {Six challenges for
  neural machine translation}.
\newblock In \emph{Proceedings of the First Workshop on Neural Machine
  Translation}, pages 28--39.

\bibitem[{Lester et~al.(2021)Lester, Al-Rfou, and Constant}]{lester2021power}
Brian Lester, Rami Al-Rfou, and Noah Constant. 2021.
\newblock \href {https://doi.org/10.18653/v1/2021.emnlp-main.243} {The power of
  scale for parameter-efficient prompt tuning}.
\newblock In \emph{Proceedings of EMNLP}, pages 3045--3059.

\bibitem[{Li and Liang(2021)}]{li2021prefix}
Xiang~Lisa Li and Percy Liang. 2021.
\newblock \href {https://doi.org/10.18653/v1/2021.acl-long.353} {Prefix-tuning:
  Optimizing continuous prompts for generation}.
\newblock In \emph{Proceedings of ACL}, pages 4582--4597.

\bibitem[{Liu et~al.(2021)Liu, Yuan, Fu, Jiang, Hayashi, and
  Neubig}]{liu2021pre}
Pengfei Liu, Weizhe Yuan, Jinlan Fu, Zhengbao Jiang, Hiroaki Hayashi, and
  Graham Neubig. 2021.
\newblock \href {https://arxiv.org/abs/2107.13586} {Pre-train, prompt, and
  predict: A systematic survey of prompting methods in natural language
  processing}.
\newblock \emph{arXiv preprint arXiv:2107.13586}.

\bibitem[{Papineni et~al.(2002)Papineni, Roukos, Ward, and
  Zhu}]{papineni2002bleu}
Kishore Papineni, Salim Roukos, Todd Ward, and WeiJing Zhu. 2002.
\newblock \href {https://doi.org/10.3115/1073083.1073135} {Bleu: A method for
  automatic evaluation of machine translation}.
\newblock In \emph{Proceedings of ACL}, pages 311--318.

\bibitem[{Post(2018)}]{post2018call}
Matt Post. 2018.
\newblock \href {https://doi.org/10.18653/v1/W18-6319} {A call for clarity in
  reporting {BLEU} scores}.
\newblock In \emph{Proceedings of the Third Conference on Machine Translation},
  pages 186--191.

\bibitem[{Radford et~al.(2018)Radford, Narasimhan, Salimans, and
  Sutskever}]{radford2018improving}
Alec Radford, Karthik Narasimhan, Tim Salimans, and Ilya Sutskever. 2018.
\newblock \href {https://openai.com/blog/language-unsupervised/} {Improving
  language understanding by generative pre-training}.
\newblock \emph{OpenAI blog}.

\bibitem[{Radford et~al.(2019)Radford, Wu, Child, Luan, Amodei, Sutskever
  et~al.}]{radford2019language}
Alec Radford, Jeffrey Wu, Rewon Child, David Luan, Dario Amodei, Ilya
  Sutskever, et~al. 2019.
\newblock \href {https://openai.com/blog/better-language-models/} {Language
  models are unsupervised multitask learners}.
\newblock \emph{OpenAI blog}.

\bibitem[{Raffel et~al.(2020)Raffel, Shazeer, Roberts, Lee, Narang, Matena,
  Zhou, Li, and Liu}]{raffel2019exploring}
Colin Raffel, Noam Shazeer, Adam Roberts, Katherine Lee, Sharan Narang, Michael
  Matena, Yanqi Zhou, Wei Li, and Peter~J Liu. 2020.
\newblock \href {http://jmlr.org/papers/v21/20-074.html} {Exploring the limits
  of transfer learning with a unified text-to-text transformer}.
\newblock \emph{Journal of Machine Learning Research}, 21(140):1--67.

\bibitem[{Shin et~al.(2020)Shin, Razeghi, Logan~IV, Wallace, and
  Singh}]{shin2020autoprompt}
Taylor Shin, Yasaman Razeghi, Robert~L Logan~IV, Eric Wallace, and Sameer
  Singh. 2020.
\newblock \href {https://doi.org/10.18653/v1/2020.emnlp-main.346}
  {{A}uto{P}rompt: {E}liciting {K}nowledge from {L}anguage {M}odels with
  {A}utomatically {G}enerated {P}rompts}.
\newblock In \emph{Proceedings of EMNLP}, pages 4222--4235.

\bibitem[{Shoeybi et~al.(2019)Shoeybi, Patwary, Puri, LeGresley, Casper, and
  Catanzaro}]{shoeybi2019megatron}
Mohammad Shoeybi, Mostofa Patwary, Raul Puri, Patrick LeGresley, Jared Casper,
  and Bryan Catanzaro. 2019.
\newblock \href {https://arxiv.org/abs/1909.08053} {{Megatron-LM: Training
  multi-billion parameter language models using model parallelism}}.
\newblock \emph{arXiv preprint arXiv:1909.08053}.

\bibitem[{Stickland et~al.(2021)Stickland, Li, and
  Ghazvininejad}]{stickland2021recipes}
Asa~Cooper Stickland, Xian Li, and Marjan Ghazvininejad. 2021.
\newblock \href {https://doi.org/10.18653/v1/2021.eacl-main.301} {Recipes for
  adapting pre-trained monolingual and multilingual models to machine
  translation}.
\newblock In \emph{Proceedings of EACL}, pages 3440--3453.

\bibitem[{Sun et~al.(2021{\natexlab{a}})Sun, Wang, Feng, Ding, Pang, Shang,
  Liu, Chen, Zhao, Lu et~al.}]{sun2021ernie}
Yu~Sun, Shuohuan Wang, Shikun Feng, Siyu Ding, Chao Pang, Junyuan Shang,
  Jiaxiang Liu, Xuyi Chen, Yanbin Zhao, Yuxiang Lu, et~al. 2021{\natexlab{a}}.
\newblock \href {https://arxiv.org/abs/2107.02137} {Ernie 3.0: Large-scale
  knowledge enhanced pre-training for language understanding and generation}.
\newblock \emph{arXiv preprint arXiv:2107.02137}.

\bibitem[{Sun et~al.(2021{\natexlab{b}})Sun, Wang, and
  Li}]{sun2021multilingual}
Zewei Sun, Mingxuan Wang, and Lei Li. 2021{\natexlab{b}}.
\newblock \href {https://doi.org/10.18653/v1/2021.findings-emnlp.233}
  {Multilingual translation via grafting pre-trained language models}.
\newblock In \emph{Findings of EMNLP}, pages 2735--2747.

\bibitem[{Sutskever et~al.(2014)Sutskever, Vinyals, and
  Le}]{sutskever2014sequence}
Ilya Sutskever, Oriol Vinyals, and Quoc~V Le. 2014.
\newblock \href
  {https://proceedings.neurips.cc/paper/2014/file/a14ac55a4f27472c5d894ec1c3c743d2-Paper.pdf}
  {Sequence to sequence learning with neural networks}.
\newblock \emph{Advances in Neural Information Processing Systems},
  27:3104--3112.

\bibitem[{Tan et~al.(2020)Tan, Zhang, Huang, Chen, Wang, Sun, Luan, and
  Liu}]{tan2020thumt}
Zhixing Tan, Jiacheng Zhang, Xuancheng Huang, Gang Chen, Shuo Wang, Maosong
  Sun, Huanbo Luan, and Yang Liu. 2020.
\newblock \href {https://aclanthology.org/2020.amta-research.11.pdf} {{THUMT:
  An Open-Source Toolkit for Neural Machine Translation}}.
\newblock In \emph{Proceedings of AMTA}, pages 116--122.

\bibitem[{Vaswani et~al.(2017)Vaswani, Shazeer, Parmar, Uszkoreit, Jones,
  Gomez, Kaiser, and Polosukhin}]{vaswani2017attention}
Ashish Vaswani, Noam Shazeer, Niki Parmar, Jakob Uszkoreit, Llion Jones,
  Aidan~N Gomez, {\L}ukasz Kaiser, and Illia Polosukhin. 2017.
\newblock \href
  {https://proceedings.neurips.cc/paper/2017/file/3f5ee243547dee91fbd053c1c4a845aa-Paper.pdf}
  {Attention is all you need}.
\newblock In \emph{Advances in Neural Information Processing Systems},
  volume~30, pages 5998--6008.

\bibitem[{Wolf et~al.(2020)Wolf, Chaumond, Debut, Sanh, Delangue, Moi, Cistac,
  Funtowicz, Davison, Shleifer et~al.}]{wolf2020transformers}
Thomas Wolf, Julien Chaumond, Lysandre Debut, Victor Sanh, Clement Delangue,
  Anthony Moi, Pierric Cistac, Morgan Funtowicz, Joe Davison, Sam Shleifer,
  et~al. 2020.
\newblock \href {https://doi.org/10.18653/v1/2020.emnlp-demos.6} {Transformers:
  State-of-the-art natural language processing}.
\newblock In \emph{Proceedings of EMNLP: System Demonstrations}, pages 38--45.

\bibitem[{Xue et~al.(2021)Xue, Constant, Roberts, Kale, Al-Rfou, Siddhant,
  Barua, and Raffel}]{xue2020mt5}
Linting Xue, Noah Constant, Adam Roberts, Mihir Kale, Rami Al-Rfou, Aditya
  Siddhant, Aditya Barua, and Colin Raffel. 2021.
\newblock \href {https://doi.org/10.18653/v1/2021.naacl-main.41} {m{T}5: A
  massively multilingual pre-trained text-to-text transformer}.
\newblock In \emph{Proceedings of NAACL}, pages 483--498.

\bibitem[{Zhang et~al.(2021)Zhang, Gu, Han, Chen, Xiao, Sun, Yao, Qi, Guan, Ke
  et~al.}]{zhang2021cpm}
Zhengyan Zhang, Yuxian Gu, Xu~Han, Shengqi Chen, Chaojun Xiao, Zhenbo Sun, Yuan
  Yao, Fanchao Qi, Jian Guan, Pei Ke, et~al. 2021.
\newblock \href {https://doi.org/10.1016/j.aiopen.2021.12.003} {{CPM-2:
  Large-scale Cost-effective Pre-trained Language Models}}.
\newblock \emph{AIOpen}, 2:216--224.

\end{thebibliography}
\bibliographystyle{acl_natbib}

\clearpage

\appendix

\section{Appendix}
\subsection{Details of Multilingual GPT}\label{sec:appendix_mgpt}
We used a multilingual GPT (mGPT)~\cite{radford2019language} model as the pre-trained language model in all our experiments. The mGPT model is trained using the Megatron-LM toolkit~\cite{shoeybi2019megatron}~\footnote{\url{https://github.com/NVIDIA/Megatron-LM}} with the default GPT-2 configuration on the mC4 dataset~\cite{xue2020mt5},~\footnote{\url{https://huggingface.co/datasets/mc4}} which contains massive web crawled data covering 101 languages. The model consists of 24 transformer layers, and the hidden size $d$ of the model is set to 1,024. We used the same tokenization and vocabulary as the mT5 model~\cite{xue2020mt5}. The vocabulary size is 250,100. The total number of parameters of the mGPT model is about 560M. We train the mGPT model on a machine with 8 RTX 3090Ti GPUs for 200K steps.

\subsection{Preprocessing and Postprocessing}\label{sec:appendix_proc}
\begin{CJK*}{UTF8}{gbsn}
We do not apply any additional preprocessing during pre-training. Preprocessing like tokenization is done automatically with the \textit{sentencepiece} program. For learning prompts, we do not apply additional preprocessing on translation tasks except Romanian-English translation task, where we use a script~\footnote{\url{https://github.com/rsennrich/wmt16-scripts/blob/master/preprocess/normalise-romanian.py}} to remove diacritics in the Romanian side. Because the mT5 tokenizer automatically uses Unicode NFKC normalization, which results in non-standard punctuation for Chinese (e.g. ``，''$\rightarrow$ ``,''). Therefore, for postprocessing, we use a rule-based method to replace non-standard punctuation with standard counterparts for Chinese.
\end{CJK*}

\subsection{Alignment Examples}\label{sec:appendix_alignments}
Table~\ref{tab:align} provides examples of induced alignments from the mGPT model without using prompts. We compute cosine similarities between target hidden keys and source hidden keys of the 15th Transformer layer of mGPT, and align the target word and the source word with the highest cosine similarity.

\begin{CJK*}{UTF8}{gbsn}
\begin{table*}[th]
\centering
\resizebox{\textwidth}{!}{ 
\begin{tabular}{l l}
\toprule
\textbf{English} & "They say there were boys around, that was not the case at all," he said. \\\midrule
\textbf{Chinese} & 他表示：“他们说周围有好几个男孩子，但事实并非如此。” \\\midrule
\textbf{Tokenized English} & \_" They \_say \_there \_were \_ boys \_around , \_that \_was \_not \_the \_case \_at \_all ," \_he \_said . \\\midrule
\textbf{Tokenized Chinese} & \_\ 他\ 表示\ :“\ 他们\ 说\ 周围\ 有\ 好\ 几个\ 男孩\ 子\ ,\ 但\ 事实\ 并非\ 如此\ 。” \\\midrule
\multirow{2}{*}{\textbf{Alignments}} & 他/\_he\ \ 表示/\_said\ \  :“/\_"\ \ 他们/They\ \ 说/\_say\ \ 周围/\_around \ \ 有/\_were \ \ 好/boys\\
                          & 几个/\_were\ \ 男孩/boys\ \ 子/boys\ \ ,/,\ \ 但/\_that\ \ 事实/\_case\ \ 并非/\_not\ \ 如此/\_all\ \ 。”/. \\\midrule\midrule
\textbf{English} & Saudi Arabia To Offer Tourist Visas For First Time, Abolish Abaya Rule \\\midrule
\textbf{Chinese} & 沙特阿拉伯首次提供旅游签证，废除阿巴亚长袍规定 \\\midrule
\textbf{Tokenized English} & \_Saudi \_Arabia \_To \_Offer \_Tourist \_Visa s \_For \_First \_Time , \_Ab olish \_A baya \_Rule \\\midrule
\textbf{Tokenized Chinese} & \_\ 沙\ 特\ 阿拉\ 伯\ 首次\ 提供\ 旅游\ 签证\ ,\ 废\ 除\ 阿\ 巴\ 亚\ 长\ 袍\ 规定 \\\midrule
\multirow{2}{*}{\textbf{Alignments}} & 沙/\_Saudi\ \ 特/\_Arabia\ \  阿拉/\_Arabia\ \ 伯/\_Arabia\ \ 首次/\_Offer\ \ 提供/\_Offer \ \ 旅游/\_Tourist\\
                          & 签证/\_Visa\ \ ,/,\ \ 废/olish\ \ 除/olish\ \ 阿/\_Saudi\ \ 巴/baya\ \ 亚/baya\ \ 长/\_Rule \ \ 袍/\_Visa\ \ 规定/\_Rule \\
\bottomrule
\end{tabular}}
\caption{Alignments induced from the mGPT model. We use ``/'' to separate Chinese and English tokens.}\label{tab:align}
\end{table*}
\end{CJK*}

\end{document}